\pdfoutput=1
\documentclass{article}
\usepackage{bm}
\usepackage{amsmath}
\usepackage{graphicx}
\usepackage{subfigure}
\usepackage{multirow}
\usepackage{url}
\usepackage{graphicx}
\usepackage{epstopdf}
\usepackage{pdfpages}
\graphicspath{{Figure/}}
\usepackage{natbib}
\usepackage{amsmath}
\usepackage{enumitem}
\setlist{nolistsep}
\usepackage{amssymb}
\usepackage{algorithm}
\usepackage{algorithmic}

\usepackage[accepted]{icml2013} 

\newtheorem{proposition}{Proposition}

\begin{document} 

\twocolumn[
\icmltitle{$\propto$SVM for Learning with Label Proportions}
\icmlauthor{Felix X. Yu$^{\dag}$}{yuxinnan@ee.columbia.edu}
\icmlauthor{Dong Liu$^{\dag}$}{dongliu@ee.columbia.edu}
\icmlauthor{Sanjiv Kumar$^{\S}$}{sanjivk@google.com}
\icmlauthor{Tony Jebara$^{\dag}$}{jebara@cs.columbia.edu}
\icmlauthor{Shih-Fu Chang$^{\dag}$}{sfchang@cs.columbia.edu}
\icmladdress{$^{\dag}$Columbia University, New York, NY 10027}
\vspace{-0.3cm}
\icmladdress{$^{\S}$Google Research, New York, NY 10011}
\vskip 0.3in
]

\begin{abstract}
We study the problem of learning with label proportions in which
the training data is provided in groups and only the proportion
of each class in each group is known. We propose a new method
called proportion-SVM, or $\propto$SVM, which explicitly models
the latent unknown instance labels together with the known group label
proportions in a large-margin framework. Unlike the existing works,
our approach avoids making restrictive assumptions about the data. The
$\propto$SVM model leads to a non-convex integer programming
problem. In order to solve it efficiently, we propose two
algorithms: one based on simple alternating optimization and the
other based on a convex relaxation. Extensive experiments on
standard datasets show that $\propto$SVM outperforms the
state-of-the-art, especially for larger group sizes.
\end{abstract}

\section{Introduction}
The problem of learning with label proportions has recently drawn
attention in the learning community~\cite{quadrianto2009estimating, rueping2010svm}. 
In this setting, the training instances are
provided as groups or ``bags''. For each bag, only the proportions of
the labels are available. The task is to learn a model to predict
labels of the individual instances.

Learning with label proportions raises multiple issues. On one hand, it enables interesting applications such as modeling voting behaviors from aggregated proportions across different demographic regions. On the other hand, the feasibility of such a learning method also raises concerns about the sensitive personal information that could potentially be leaked simply by observing label proportions.

\vspace{-0.05cm}
To address this learning setting, this article explicitly models the unknown instance labels as latent variables. This alleviates the need for making restrictive assumptions on the data, either parametric or generative. We introduce a large-margin framework called proportion-SVM, or $\propto$SVM\footnote{$\propto$ is the symbol for ``proportional-to''.}, which jointly optimizes over the unknown
 instance labels and the known label proportions (Section
 \ref{sec:pSVM}). In order to solve $\propto$SVM efficiently, we propose two
 algorithms - one based on simple alternating optimization (Section
 \ref{sec:alter}), and the other based on a convex relaxation (Section
 \ref{sec:conv}). We show that our approach outperforms the existing methods for various
 datasets and settings (Section \ref{sec:exp}). The gains are especially
 higher for more challenging settings when the bag size is large.
 
\vspace{-0.1cm}
\section{Related Works}
\label{sec:related}
\textbf{MeanMap:}
\citet{quadrianto2009estimating} proposed a theoretically sound
method to estimate the mean of each class using the mean of each bag
and the label proportions. These estimates are then used in a
conditional exponential model to maximize the log likelihood.
The key assumption in MeanMap is that the class-conditional distribution of data
 is independent of the bags.  Unfortunately, this
assumption does not hold for many real world applications. For
example, in modeling voting behaviors, in which the bags are different
demographic regions, the data distribution can be highly dependent on
the bags.

\textbf{Inverse Calibration (InvCal):} \citet{rueping2010svm}
proposed treating the mean of each bag as a ``super-instance'', which
was assumed to have a soft label corresponding to the label
proportion. The
``super-instances'' can be poor in representing the properties of the
bags. Our work also utilizes a large-margin
framework, but we explicitly model the instance labels. Section
\ref{subsec:invcal} gives a detailed comparison with InvCal.

Figure~\ref{fig:bag} provides a toy example to highlight the problems with
MeanMap and InvCal, which are the state-of-the-art methods.
 
\textbf{Related Learning Settings:}
In semi-supervised learning, \citet{mann2007simple} and \citet{bellare2009alternating} 
used an expectation regularization term to encourage model 
predictions on the unlabeled data to match the given proportions. 
Similar ideas were also studied in the generalized 
regularization method \cite{gillenwater2011posterior}.
\citet{li2009semi} proposed a variant of semi-supervised SVM 
to incorporate the label mean of the unlabeled data. 
Unlike semi-supervised learning, the learning setting
we are considering requires no instance labels for training.

As an extension to multiple-instance learning, \citet{Kuck05} 
designed a hierarchical probabilistic model to generate consistent label proportions. 
Besides the inefficiency in optimization, the method was shown to be inferior to MeanMap \cite{quadrianto2009estimating}. Similar ideas have also been 
studied by \citet{chen2006learning} and \citet{musicant2007supervised}.

\citet{stolpe2011learning} proposed an evolutionary strategy 
paired with a labeling heuristic for clustering with label proportions. 
Different from clustering, the proposed $\propto$SVM framework 
jointly optimizes the latent instance labels and a large-margin classification model.
The $\propto$SVM formulation is related to large-margin 
clustering \cite{xu2004maximum}, with an additional objective 
to utilize the label proportions. Specifically, the convex 
relaxation method we used is inspired by the works 
of \citet{li2009semi} and \citet{xu2004maximum}.

\section{The $\propto$SVM Framework}
\label{sec:pSVM}

\subsection{Learning Setting}

We consider a binary learning setting as follows.
The training set $\{\mathbf{x}_i\}_{i=1}^N$ is given in the form of $K$ bags,

\vspace{-0.2cm}
\begin{small}
\begin{equation}
\{\mathbf{x}_i | i \in \mathcal{B}_k\}_{k=1}^K, \quad \cup_{k=1}^K \mathcal{B}_k = \{1   \cdots   N\}. 
\end{equation}
\end{small}
\vspace{-0.5cm}

In this paper, we assume that the bags are disjoint, \emph{i.e.}, $\mathcal{B}_k \cap \mathcal{B}_l = \emptyset$, $\forall k \neq l$.
The $k$-th bag is with label proportion $p_k$:

\vspace{-0.4cm}
\begin{small}
\begin{equation}
\forall_{k=1}^K, \quad p_k := \frac{|\{i | i \in \mathcal{B}_k,  y_i^* = 1\}|}{|\mathcal{B}_k|},
\end{equation}
\end{small}
\vspace{-0.62cm}

in which $y_i ^ * \in \{1, -1\}$ denotes the \emph{unknown} ground-truth label of $\mathbf{x}_i$, $\forall_{i=1}^N$.
We use $f(x) = \text{sign}(\mathbf{w}^T \varphi(\mathbf{x}) + b)$ for predicting the binary label of an instance $\mathbf{x}$, where $\varphi(\cdot)$ is a map of the input data.

\subsection{Formulation}
We explicitly model the unknown instance labels as $\mathbf{y} = (y_1, \cdots, y_N )^T$, in which $y_i \in \{-1, 1\}$ denotes the unknown label of $\mathbf{x}_i$, $\forall_{i=1}^N$. Thus the label proportion of the $k$-th bag can be straightforwardly modeled as

\begin{small}
\vspace{-0.25cm}
\begin{equation}
\tilde{p}_k(\mathbf{y}) = \frac{|\{i | i \in \mathcal{B}_k,  y_i = 1\}|}{|\mathcal{B}_k|} =  \frac{\sum_{i \in \mathcal{B}_k}{y_i}}{2|\mathcal{B}_k|} + \frac{1}{2}.
\label{eq:tilde_p_k}
\end{equation}
\vspace{-0.4cm}
\end{small}

We formulate the $\propto$SVM under the large-margin framework as below.

\begin{small}
\vspace{-0.5cm}
\begin{align}
\label{eq:pSVM_2}
\min_{\mathbf{y}, \mathbf{w}, b} \quad \!\! \!\!  & {\frac{1}{2} \mathbf{w}^T \! \mathbf{w}} \! + \! C\! \sum_{i = 1} ^ N \! L(y_i, \mathbf{w}^T \! \varphi(\mathbf{x}_i) \!+\! b)  
\! +\! C_p \! \sum_{k=1}^K \! L_p\left(\tilde{p}_k(\mathbf{y}), p_k \right) \nonumber \\
\text{s.t.}   & \quad \!\! \forall_{i=1}^N, \quad \!\!\!  y_i \in \{-1, 1\}, 
\end{align}
\vspace{-0.5cm}
\end{small}

in which $L(\cdot) \geq 0$ is the loss function for classic supervised learning.
$L_p(\cdot) \geq 0$ is a function to penalize the difference between the true label proportion and the estimated label proportion based on $\mathbf{y}$. The task is to simultaneously optimize the labels $\mathbf{y}$ and the model parameters $\mathbf{w}$ and $b$.

The above formulation permits using different loss functions for $L(\cdot)$ and $L_p(\cdot)$. 
One can also add weights for different bags.
Throughout this paper, we consider $L(\cdot)$ as the hinge loss, which is widely used for large-margin learning:
$L(y_i, \mathbf{w}^T \varphi(\mathbf{x}_i) + b) = \max \left(0, 1-y_i (\mathbf{w}^T \varphi(\mathbf{x}_i) + b) \right)$.
The algorithms in Section \ref{sec:alter} and Section \ref{sec:conv} can be easily generalized to different $L(\cdot)$.

Compared to \cite{rueping2010svm, quadrianto2009estimating}, $\propto$SVM
requires no restrictive assumptions on the data. In fact, in the special case
where no label proportions are provided, $\propto$SVM becomes large-margin
clustering \cite{xu2004maximum, li2009semi}, whose solution depends
only on the data distribution.  
$\propto$SVM can naturally incorporate any amount of
supervised data without modification. The labels for such instances will
be observed variables instead of being hidden.
$\propto$SVM can be easily extended
to the multi-class case, similar to \cite{keerthis2013}.

\subsection{Connections to InvCal}
\label{subsec:invcal}
As stated in Section \ref{sec:related}, the Inverse Calibration method (InvCal) (\citealt{rueping2010svm}) treats the mean of each bag as a ``super-instance'', which is assumed to have a soft label corresponding to the label proportion.
It is formulated as below.

\vspace*{-0.5cm}
\begin{small}
\begin{align}
\label{eq:rueping_svm}
\min_{\mathbf{w},b, \bm{\xi}, \bm{\xi}^*} \quad & \frac{1}{2} \mathbf{w}^T\mathbf{w} + C_p\sum_{k = 1}^K (\xi_k + \xi_k^*) \\ \nonumber
\forall_{k = 1}^{K},  \quad & \xi_k \geq 0, \quad \xi_k^* \geq 0 \\ \nonumber
\forall_{k = 1}^{K}, \quad & \mathbf{w}^T \bm{m}_k + b \geq -\log(\frac{1}{p_k} - 1) - \epsilon_k - \xi_k \\  \nonumber
\quad &  \mathbf{w}^T \bm{m}_k + b \leq -\log(\frac{1}{p_k} - 1) + \epsilon_k + \xi_k^*, \nonumber
\end{align}
\vspace{-0.5cm}
\end{small}

in which the $k$-th bag mean is  $\bm{m}_k = \frac{1}{
  |\mathcal{B}_k|} \sum_{i \in \mathcal{B}_k} \varphi(\mathbf{x}_i)$,
$\forall_{k=1}^K$.  Unlike $\propto$SVM, the proportion of the $k$-th
bag is modeled on top of this ``super-instance'' $\bm{m}_k$ as:

\begin{small}
\vspace{-0.5cm}
\begin{align}
\label{eq:rueping}
q_k := \left( 1+\exp{\left( - \mathbf{w}^T  \bm{m}_k  + b \right)} \right)^{-1}.
\end{align}
\vspace{-0.4cm}
\end{small}

The second term of the objective function (\ref{eq:rueping_svm}) tries
to impose $q_k \approx p_k$, $\forall_{k=1}^K$, albeit in an inverse
way.

\begin{figure}[t]
  \centering
  \vspace{-0.3cm}
  \subfigure[Bag 1, with $p_1 = 0.6$]{\includegraphics[width = 3.9cm]{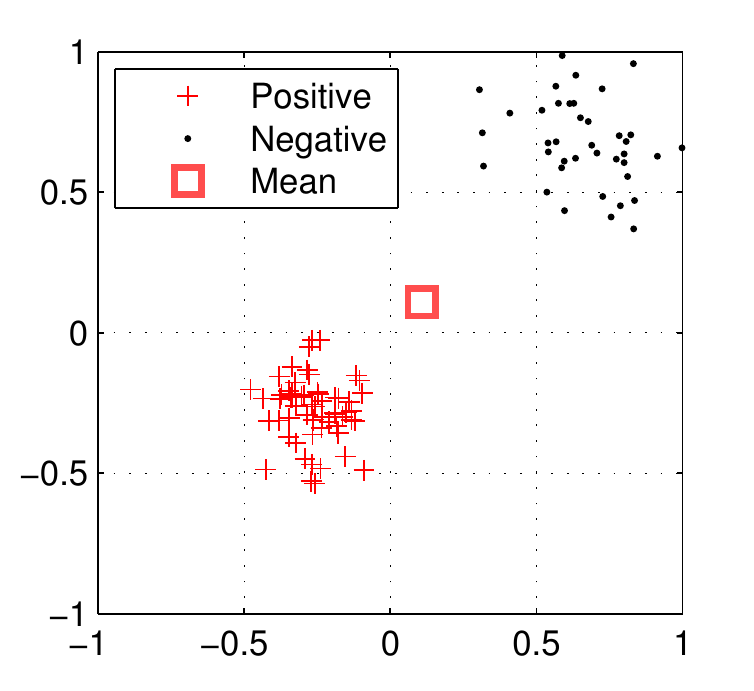}} \quad
  \subfigure[Bag 2, with $p_2 = 0.4$]{\includegraphics[width = 3.9cm]{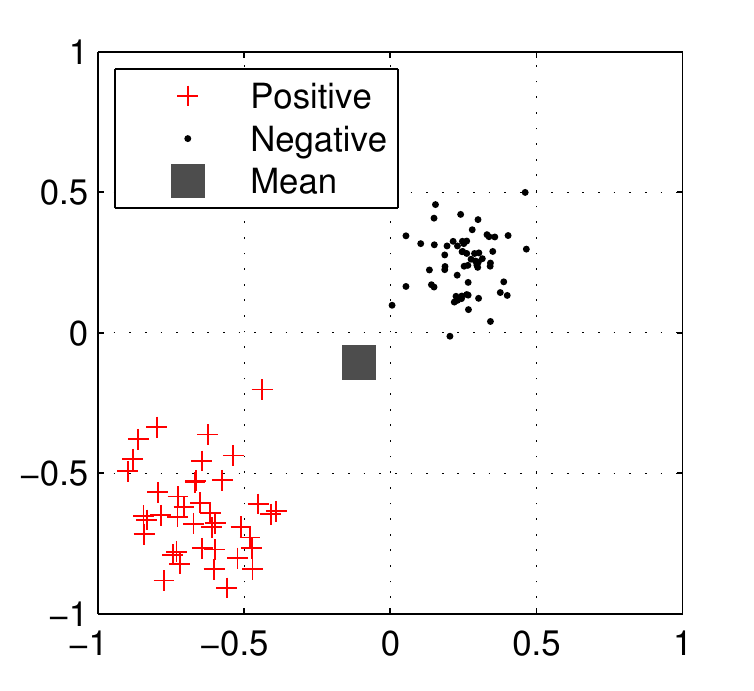}} \\
  \vspace{-0.2cm}
  \subfigure[MeanMap and InvCal with 0\% accuracy.]{\includegraphics[width = 3.9cm]{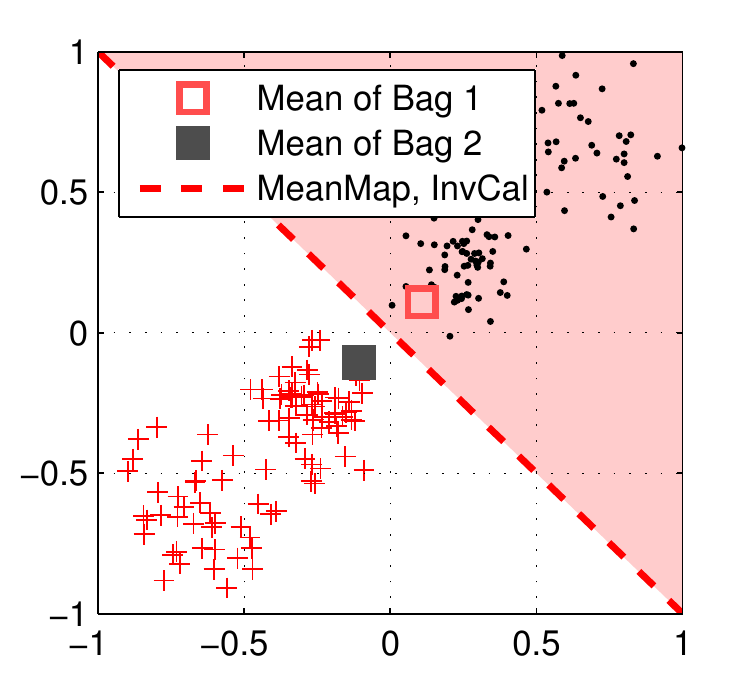}} \quad
  \subfigure[$\propto$SVM with 100\% accuracy.]{\includegraphics[width = 3.9cm]{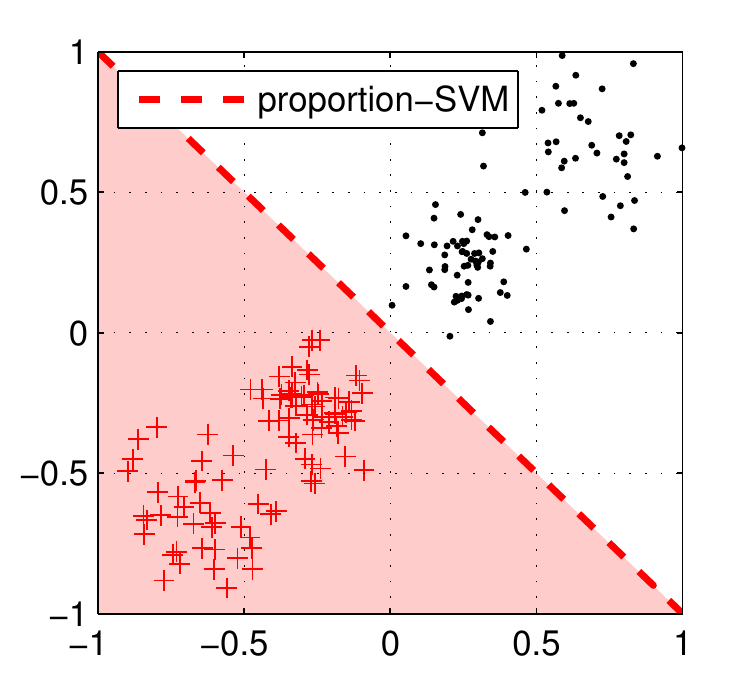}}
\caption{An example of learning with two bags to illustrate the drawbacks of the existing methods. (a) Data of bag 1. (b) Data of bag 2. (c) Learned separating hyperplanes of MeanMap and InvCal. (d) Learned separating hyperplane of $\propto$SVM (either alter-$\propto$SVM or conv-$\propto$SVM). More details are given in Section \ref{sub:toy}. Note that the algorithms do not have access to the individual instance labels.}
\label{fig:bag}
\vspace{-0.3cm}
\end{figure}

Though InvCal is shown to outperform other alternatives, including
MeanMap \cite{quadrianto2009estimating} and several simple
large-margin heuristics, it has a crucial limitation.  Note that
(\ref{eq:rueping}) is not a good way of measuring the proportion
predicted by the model, especially when the data has high variance,
or the data distribution is dependent on the bags. In our formulation
(\ref{eq:pSVM_2}), by explicitly modeling the unknown instance labels
$\mathbf{y}$, the label proportion can be directly modeled as
$\tilde{p}_k(\mathbf{y})$ given in (\ref{eq:tilde_p_k}). The advantage of our
method is illustrated in a toy experiment shown in Figure
\ref{fig:bag} (for details see Section \ref{sub:toy}).

\subsection{Difficulties in Solving $\propto$SVM}
The $\propto$SVM formulation is fairly intuitive and straightforward. It is, however, a non-convex integer programming problem, which is NP-hard. Therefore, one key issue lies in how to find an efficient algorithm to solve it approximately.
In this paper, we provide two solutions: a simple alternating optimization method (Section \ref{sec:alter}), and a convex relaxation method (Section \ref{sec:conv}).

\section{The alter-$\propto$SVM Algorithm}
\label{sec:alter}

In $\propto$SVM, the unknown instance labels $\mathbf{y}$ can be seen as a bridge between supervised learning loss and label proportion loss. Therefore, one natural way for solving (\ref{eq:pSVM_2}) is via alternating optimization as,

\begin{itemize}
\item For a fixed $\mathbf{y}$, the optimization of (\ref{eq:pSVM_2}) $w.r.t$ $\mathbf{w}$ and $b$ becomes a classic SVM problem.
\item When $\mathbf{w}$ and $b$ are fixed, the problem becomes:
\end{itemize}

\vspace{-0.4cm}
\begin{small}
\begin{align}
\label{eq:solving_y}
\min_{\mathbf{y}} \quad  & \sum_{i = 1} ^ N L(y_i, \mathbf{w}^T \varphi(\mathbf{x}_i) + b)+ \frac{C_p}{C} \sum_{k=1}^K L_p \left(\tilde{p}_k(\mathbf{y}), p_k \right) \nonumber   \\
\text{s.t.} \quad &  \forall_{i=1}^N, \quad y_i \in \{1,-1\}. 
\end{align}
\vspace{-0.4cm}
\end{small}

We show that the second step above can be solved efficiently.
Because the influence of each bag $\{y_i | i\in \mathcal{B}_k \}$, $\forall_{k=1}^K$ on the objective is independent, we can optimize (\ref{eq:solving_y})  on each bag separately. In particular, solving $\{y_i | i \in \mathcal{B}_k\}$ yields the following optimization problem:

\begin{small}
\vspace{-0.4cm}
\begin{align}
\label{eq:prop_1}
\min_{ \{y_i | i \in \mathcal{B}_k\} } \quad & \sum_{i \in \mathcal{B}_k} L(y_i, \mathbf{w}^T \varphi(\mathbf{x}_i) + b) + \frac{C_p}{C}  L_p \left( \tilde{p}_k(\mathbf{y}), p_k \right) \nonumber  \\
\text{s.t.} \quad &  \forall i \in \mathcal{B}_k,\quad y_i \in \{1,-1\}. 
\end{align}
\vspace{-0.5cm}
\end{small}

\begin{proposition}
\label{prop:inner}
For a fixed $\tilde{p}_k(\mathbf{y}) = \theta$, (\ref{eq:prop_1}) can be optimally solved by the steps below.
\end{proposition}
\begin{itemize}[leftmargin=0cm,itemindent=0.5cm,labelwidth=\itemindent,labelsep=0cm,align=left]
\item Initialize $y_i = -1$, $i \in \mathcal{B}_k$. The optimal solution can be obtained by flipping the signs as below.
\item By flipping the sign of $y_i$, $i \in \mathcal{B}_k$, suppose the reduction of the first term in (\ref{eq:prop_1}) is $\delta_i$. Sort $\delta_i$, $i \in \mathcal{B}_k$. Then flip the signs of the top-$R$ $y_i$'s which have the highest reduction $\delta_i$. 
$R = \theta |\mathcal{B}_k|$.
\end{itemize}

For bag $\mathcal{B}_k$, we only need to sort the corresponding $\delta_i$, $i \in \mathcal{B}_k$ once. Sorting takes $\mathcal{O}(|\mathcal{B}_k| \log(|\mathcal{B}_k|))$ time.
After that, for each $\theta \in \{0, \frac{1}{|\mathcal{B}_k|}, \frac{2}{|\mathcal{B}_k|},   \cdots  , 1\}$, the optimal solution can be computed incrementally, each taking $\mathcal{O}(1)$ time. We then pick the solution with the smallest objective value, yielding the optimal solution of (\ref{eq:prop_1}).

\begin{proposition}
\label{prop:y}
Following the above steps, (\ref{eq:solving_y}) can be solved in $\mathcal{O}(N \log (J))$ time, $J = \max_{k = 1  \cdots  K}|\mathcal{B}_k|$.
\end{proposition}
\vspace{-0.1cm}

The proofs of the above propositions are given in the supplementary material.
\vspace{-0.05cm}

By alternating between solving ($\mathbf{w}$, $b$) and $\mathbf{y}$, the objective is guaranteed to converge. 
This is due to the fact that the objective function is lower bounded, and non-increasing.
In practice, we terminate the procedure when the objective no longer decreases (or if its decrease is smaller than a threshold). Empirically, the alternating optimization typically terminates fast within tens of
iterations, but one obvious problem is the possibility of local solutions.

\begin{algorithm}[t]
\begin{small}
   \caption{alter-$\propto$SVM}
   \label{alg:alt}
\begin{algorithmic}
   \STATE Randomly initialize $y_i \in \{-1, 1\}$, $\forall_{i=1}^N$. $C^* = 10^{-5}C$.
   \WHILE{$C^* < C$}
   \STATE{$C^* = \min\{(1+ \Delta) C^*, C\}$}
   \REPEAT
   \STATE{Fix $\mathbf{y}$ to solve $\mathbf{w}$ and $b$.}
   \STATE{Fix $\mathbf{w}$ and $b$ to solve $\mathbf{y}$ (Eq. (\ref{eq:solving_y}) with $C \leftarrow C^*$}).
   \UNTIL{The decrease of the objective is smaller than a threshold ($10^{-4}$)}
   \ENDWHILE
\end{algorithmic}
\end{small}
\end{algorithm}

To alleviate this problem, similar to T-SVM
\cite{joachims1999transductive, chapelle2008optimization}, 
the proposed alter-$\propto$SVM algorithm
(Algorithm \ref{alg:alt}) takes an additional annealing loop to
gradually increase $C$. 
Because the nonconvexity of the objective function mainly comes from the second term of (\ref{eq:pSVM_2}), 
the annealing can be seen as a ``smoothing'' step to protect the algorithm from sub-optimal solutions. 
Following \cite{chapelle2008optimization}, we set $\Delta = 0.5$ in Algorithm \ref{alg:alt} throughout this work.
The convergence and annealing are further discussed in the supplementary material.

In the implementation of alter-$\propto$SVM, we consider $L_p(\cdot)$ as the absolute loss: $L_p(\tilde{p}_k(\mathbf{y}), p_k) = |\tilde{p}_k(\mathbf{y}) - p_k|$. 
Empirically, 
each alter-$\propto$SVM loop given an annealing value $C^*$ terminates within a few iterations.
From Proposition \ref{prop:y}, optimizing $\mathbf{y}$ has linear complexity 
in $N$ (when $J$ is small). Therefore the overall complexity of the
algorithm depends on the SVM solver.
Specifically, when linear SVM is used \cite{joachims2006training}, alter-$\propto$SVM has linear complexity.
In practice, to further alleviate the influence of the local solutions, similar to clustering, \emph{e.g.}, kmeans, we repeat alter-$\propto$SVM multiple times by randomly initializing $\mathbf{y}$, and then picking the solution with the smallest objective value.

\vspace{-0.1cm}

\section{The conv-$\propto$SVM Algorithm}
\label{sec:conv}

In this section, we show that with proper relaxation of the $\propto$SVM formulation (\ref{eq:pSVM_2}), the objective function can be transformed to a convex function of $\bm{M} := \mathbf{y}\mathbf{y}^T$.
We then relax the solution space of $\bm{M}$ to its convex hull, leading to a convex optimization problem of $\bm{M}$. 
The conv-$\propto$SVM algortihm is proposed to solve the relaxed problem. 
Unlike alter-$\propto$SVM, conv-$\propto$SVM does not require multiple initializations. 
This method is motivated by the techniques used in large-margin clustering \cite{li2009tighter, xu2004maximum}.

\subsection{Convex Relaxation}
We change the label proportion term in the objective function (\ref{eq:pSVM_2}) as a constraint $\mathbf{y} \in \mathcal{Y}$, and we drop the bias term $b$\footnote{If the bias term is not dropped, there will be constraint $\bm{\alpha}^T \mathbf{y} = 0$ in the dual, leading to non-convexity. Such difficulty has also been discussed in \cite{xu2004maximum}. Fortunately, the effect of removing the bias term can be alleviated by zero-centering the data or augmenting the feature vector with an additional dimension with value $1$.}. Then, (\ref{eq:pSVM_2}) is rewritten as:

\begin{small}
\vspace{-0.5cm}
\begin{align}
\label{eq:convex_1}
&\min_{\mathbf{y} \in \mathcal{Y}}\min_{\mathbf{w}} \quad {\frac{1}{2} \mathbf{w}^T \mathbf{w}} + C \sum_{i = 1} ^ N L(y_i, \mathbf{w}^T \varphi(\mathbf{x}_i)) \\
&\mathcal{Y} = \Big\{\mathbf{y}\big| |\tilde{p}_k(\mathbf{y}) - p_k | \leq \epsilon, y_i \in \{-1,1\}, \forall_{k=1}^K \Big\}, \nonumber
\end{align}
\vspace{-0.5cm}
\end{small}

in which $\epsilon$ controls the compatibility of the label proportions. 
The constraint $\mathbf{y} \in \mathcal{Y}$ can be seen as a special loss function:

\begin{small}
\vspace{-0.6cm}
\begin{align}
L_p(\tilde{p}_k(\mathbf{y}), p_k) =
 \begin{cases}
  0, & \text{if } |\tilde{p}_k(\mathbf{y}) - p_k| < \epsilon,  \\
  \infty, & \text{otherwise}.
\end{cases}
\end{align}
\vspace{-0.4cm}
\end{small}

We then write the inner problem of (\ref{eq:convex_1}) as its dual:

\begin{small}
\vspace{-0.4cm}
\begin{align}
\label{eq:convex_2}
\min_{\mathbf{y} \in \mathcal{Y}}\max_{\bm{\alpha} \in \mathcal{A}} -\frac{1}{2} \bm{\alpha}^T \left(
\bm{\mathcal{K}} \odot \mathbf{y}\mathbf{y}^T \right) \bm{\alpha} + \bm{\alpha}^T \bm{1},
\end{align}
\vspace{-0.4cm}
\end{small}

in which $\bm{\alpha} \in \mathbb{R}^N$, $\odot$ denotes pointwise-multiplication, $\bm{\mathcal{K}}$ is the kernel matrix with $\mathcal{K}_{ij} = \varphi(\mathbf{x}_i)^T\varphi(\mathbf{x}_j)$, $\forall_{i,j = 1}^N$, and $\mathcal{A} = \{\bm{\alpha} | 0 \leq \bm{\alpha} \leq C \}$.

The objective in (\ref{eq:convex_2}) is non-convex in $\mathbf{y}$, but convex in $\bm{M} := \mathbf{y}\mathbf{y}^T$. So, following \cite{li2009tighter, xu2004maximum}, we instead solve the optimal $\bm{M}$. However, the feasible space of $\bm{M}$ is

\begin{small}
\vspace{-0.3cm}
\begin{equation}
\mathcal{M}_0 = \{\mathbf{y} \mathbf{y}^T | \mathbf{y} \in \mathcal{Y} \},
\end{equation}
\vspace{-0.5cm}
\end{small}

which is a non-convex set.  In order to get a convex optimization problem, we relax $\mathcal{M}_0$ to its convex hull, the tightest convex relaxation of $\mathcal{M}_0$:

\begin{small}
\vspace{-0.3cm}
\begin{equation}
\mathcal{M} =  \Big\{ \sum_{\mathbf{y} \in \mathcal{Y}}
\mu_{(\mathbf{y})} \mathbf{y} \mathbf{y}^T   \Big| \bm{\mu} \in \mathcal{U} \Big\},
\end{equation}
\vspace{-0.3cm}
\end{small}

in which $\mathcal{U} = \{\bm{\mu} | \sum_{\mathbf{y} \in \mathcal{Y}} \mu_{(\mathbf{y})} = 1, \mu_{(\mathbf{y})} \geq 0\}$.

Thus solving the relaxed $\bm{M}$ is identical to finding $\bm{\mu}$:

\vspace*{-0.4cm}
\begin{small}
\begin{align}
\label{eq:convex_3}
\min_{\bm{\mu} \in \mathcal{U}} \max_{\bm{\alpha} \in \mathcal{A}} -\frac{1}{2} \bm{\alpha}^T \left( \sum_{\mathbf{y} \in \mathcal{Y}} \mu_{(\mathbf{y})}
\bm{\mathcal{K}} \odot \mathbf{y}\mathbf{y}^T  \right) \bm{\alpha} + \bm{\alpha}^T \bm{1}.
\end{align}
\vspace{-0.3cm}
\end{small}

(\ref{eq:convex_3}) can be seen as Multiple Kernel Learning (MKL) \cite{bach2004multiple}, which is a widely studied problem.
However, because $|\mathcal{Y}|$ is very large, it is not tractable to solve (\ref{eq:convex_3}) directly.

\subsection{Cutting Plane Training}
Fortunately, we can assume that at optimality only a small number of $\mathbf{y}$'s are active in (\ref{eq:convex_3}).
Define $\mathcal{Y}_{active} \subset \mathcal{Y}$ as the set containing all the active $\mathbf{y}$'s.
We show that $\mathbf{y} \in \mathcal{Y}_{active}$ can be incrementally found by the cutting plane method.

Because the objective function of (\ref{eq:convex_3}) is convex in $\bm{\mu}$, and concave in $\bm{\alpha}$,
it is equivalent to \cite{fan1953minimax},

\begin{small}
\vspace{-0.4cm}
\begin{align}
\max_{\bm{\alpha} \in \mathcal{A}} \min_{\bm{\mu} \in \mathcal{U}}  -\frac{1}{2} \bm{\alpha}^T \left( \sum_{\mathbf{y} \in \mathcal{Y}} \mu_{(\mathbf{y})}
\bm{\mathcal{K}} \odot \mathbf{y}\mathbf{y}^T \right) \bm{\alpha} + \bm{\alpha}^T \bm{1}.
\end{align}
\vspace{-0.4cm}
\end{small}

It is easy to verify that the above is equivalent to:

\begin{small}
\vspace{-0.4cm}
\begin{align}
\label{eq:convex_4}
\max_{\bm{\alpha} \in \mathcal{A}, \beta} \quad &  -\beta \\
\text{s.t.} \quad & \beta \geq  \frac{1}{2} \bm{\alpha}^T \left(
\bm{\mathcal{K}} \odot \mathbf{y}\mathbf{y}^T
 \right) \bm{\alpha} + \bm{\alpha}^T \bm{1}, \forall \mathbf{y} \in \mathcal{Y} . \nonumber                                
\end{align}
\vspace{-0.4cm}
\end{small}

This form enables us to apply the cutting plane method \cite{kelley1960cutting} to incrementally include the most violated $\mathbf{y}$ into $\mathcal{Y}_{active}$, and then solve the MKL problem, (\ref{eq:convex_3}) with $\mathcal{Y}$ replaced as $\mathcal{Y}_{active}$. The above can be repeated until no violated $\mathbf{y}$ exists.

In the cutting plane training, the critical step is to obtain the most violated $\mathbf{y} \in \mathcal{Y}$:

\begin{small}
\vspace{-0.3cm}
\begin{equation}
\arg\max_{\mathbf{y} \in \mathcal{Y}}\frac{1}{2}\bm{\alpha}^T \left(\bm{\mathcal{K}} \odot \mathbf{y}\mathbf{y}^T \right) \bm{\alpha} + \bm{\alpha}^T \bm{1},
\end{equation}
\vspace{-0.3cm}
\end{small}

which is equivalent to

\begin{small}
\vspace{-0.4cm}
\begin{equation}
\label{eq:concave}
\arg\max_{\mathbf{y} \in \mathcal{Y}} \sum_{i,j = 1}^N \alpha_i \alpha_j y_i y_j \varphi(\mathbf{x}_i)^T \varphi(\mathbf{x}_j).
\end{equation}
\vspace{-0.3cm}
\end{small}

This is a 0/1 concave QP, for which there exists no efficient
solution. However, instead of finding the most violated constraint, if
we find any violated constraint $\mathbf{y}$, the objective
function still decreases. We therefore relax the objective in
(\ref{eq:concave}), which can be solved efficiently. Note that the objective of (\ref{eq:concave}) is equivalent to a $\ell_2$ norm  $\sum_{i = 1}^N \parallel \alpha_i  y_i \varphi(\mathbf{x}_i) \parallel_2$. Following \cite{li2009tighter}, we approximate it as the $\ell_\infty$ norm:

\begin{small}
\vspace{-0.5cm}
\begin{align}
\label{eq:inf}
 \sum_{i = 1}^N \parallel \alpha_i  y_i \varphi(\mathbf{x}_i) \parallel_\infty \equiv \max_{j = 1 \cdots  d}\left| \sum_{i=1}^N \alpha_i y_i x_i^{(j)} \right|,
\end{align}
\vspace{-0.4cm}
\end{small}

in which $x_i^{(j)}$ is the $j$-th dimension of the $i$-th feature vector. These can be obtained by eigendecomposition of the kernel matrix $\bm{\mathcal{K}}$, when a nonlinear kernel is used. The computational complexity is $\mathcal{O}(dN^2)$. In practice, we choose $d$ such that 90\% of the variance is preserved. We further rewrite (\ref{eq:inf}) as:

\begin{small}
\vspace{-0.3cm}
\begin{align}
& \max_{j = 1 \cdots d}  \max \left( \sum_{i=1}^N \alpha_i y_i x_i^{(j)},  -\sum_{i=1}^N \alpha_i y_i x_i^{(j)} \right) \\
&=  \max_{j = 1 \cdots d} \max \left( \sum\limits_{k = 1}^K  \sum\limits_{i \in \mathcal{B}_k} \alpha_i y_i x_i^{(j)} ,  \sum\limits_{k = 1}^K \sum\limits_{i \in \mathcal{B}_k} -\alpha_i y_i x_i^{(j)} \right). \nonumber
\label{eq:convex_vio}
\end{align}
\vspace{-0.3cm}
\end{small}

Therefore the approximation from (\ref{eq:concave}) to (\ref{eq:inf}) enables us to consider each dimension and each bag separately.
For the $j$-th dimension, and the $k$-th bag, we only need to solve two sub-problems ${\max_{\mathbf{y} \in \mathcal{Y}} \sum_{i \in \mathcal{B}_k} \alpha_i y_i x_i^{(j)}}$, and $\max_{\mathbf{y} \in \mathcal{Y}} -\sum_{i \in \mathcal{B}_k} \alpha_i y_i x_i^{(j)}$. The former, as an example, can be written as

\begin{small}
\vspace{-0.3cm}
\begin{align}
\min_{\{y_i | i \in \mathcal{B}_k\} } \sum_{i \in \mathcal{B}_k} \left(-\alpha_i x_i^{(j)}\right)y_i, \quad |\tilde{p}_k(\mathbf{y})-p_k| \leq \epsilon.
\end{align}
\vspace{-0.3cm}
\end{small}

This can be solved in the same way as (\ref{eq:prop_1}), 
which takes $\mathcal{O}(|\mathcal{B}_k|\log|\mathcal{B}_k|)$ time.
Because we have $d$ dimensions, similar to Proposition \ref{prop:y}, 
one can show that:

\begin{proposition}
\label{prop:conv_2}
(\ref{eq:concave}) with the $\ell_2$ norm approximated as the 
$\ell_{\infty}$ norm can be solved in $\mathcal{O}(dN\log(J))$ time, 
$J = \max_{k= 1 \cdots K} |\mathcal{B}_k|$.
\end{proposition}

\begin{algorithm}[t]
\begin{small}
   \caption{conv-$\propto$SVM}
   \label{alg:conv}
\begin{algorithmic}
   \STATE Initialize $\alpha_i = 1/N$, $\forall_{i=1}^N$. $\mathcal{Y}_{active} = \emptyset$. Output: $\bm{M} \in \mathcal{M}$
   \REPEAT
   \STATE{Compute $\mathbf{y} \in \mathcal{Y}$ based on (\ref{eq:concave}) $-$ (\ref{eq:convex_vio}).}
   \STATE{ $\mathcal{Y}_{active} \leftarrow \mathcal{Y}_{active} \cup \{\mathbf{y}\}$.}
   \STATE{Solve the MKL problem in (\ref{eq:convex_3}) with $\mathcal{Y}_{active}$ to get $ \mu_{(\mathbf{y})}$, $\mathbf{y}\in \mathcal{Y}_{active}$.}
   \UNTIL{The decrease of the objective is smaller than a threshold ($10^{-4}$)}
\end{algorithmic}
\end{small}
\end{algorithm}

\subsection{The Algorithm}

The overall algorithm, called conv-$\propto$SVM, is shown in Algorithm \ref{alg:conv}.
Following \cite{li2009tighter}, we use an adapted SimpleMKL algorithm 
\cite{rakotomamonjy2008simplemkl} to solve the MKL problem.

As an additional step, we need to recover $\mathbf{y}$ from $\bm{M}$. 
This is achieved by rank-1 approximation of $\bm{M}$ (as $\mathbf{y}\mathbf{y}^T$)\footnote{Note that $\mathbf{y}\mathbf{y}^T = (-\mathbf{y})(-\mathbf{y})^T$. This 
ambiguity can be resolved by validation on the training bags.}. 
Because of the convex relaxation, the computed $\mathbf{y}$ is not binary.
However, we can use the real-valued $\mathbf{y}$ directly in our prediction model (with dual):

\begin{small}
\vspace{-0.5cm}
\begin{equation}
f(\mathbf{x}) = \text{sign}\left( \sum_{i=1}^N \alpha_i y_i \varphi(\mathbf{x}_i)^T \varphi(\mathbf{x}) \right).
\end{equation}
\vspace{-0.5cm}
\end{small}

Similar to alter-$\propto$SVM, the objective of conv-$\propto$SVM is 
guaranteed to converge. In practice, we terminate the algorithm when the decrease of the objective
is smaller than a threshold. 
Typically the SimpleMKL converges in less than 5 iterations, 
and conv-$\propto$SVM terminates in less than 10 iterations.
The SimpleMKL takes $\mathcal{O}(N^2)$ (computing the gradient) time, or 
the complexity of SVM, whichever is higher. Recovering 
$\mathbf{y}$ takes $\mathcal{O}(N^2)$ time and computing eigendecomposition 
with the first $d$ singular values takes $\mathcal{O}(dN^2)$ time.

\section{Experiments}
\label{sec:exp}

MeanMap \cite{quadrianto2009estimating} was shown to outperform alternatives including kernel density estimation, discriminative sorting and MCMC \cite{Kuck05}. InvCal \cite{rueping2010svm} was shown to outperform MeanMap and several large-margin alternatives. Therefore,
in the experiments, we only compare our approach with MeanMap and InvCal.

\subsection{A Toy Experiment}
\label{sub:toy}
To  visually demonstrate the advantage of our approach, we first show an experiment on a toy dataset with two bags.
Figure \ref{fig:bag} (a) and (b) show the data of the two bags, and Figure \ref{fig:bag} (c) and (d) show the learned separating hyperplanes from different methods. Linear kernel is used in this experiment.
For this specific dataset, the restrictive data assumptions of MeanMap and InvCal do not hold: the mean of the first bag (60\% positive) is on the ``negative side'', whereas, the mean of the second bag (40\% positive) is on the ``positive side''. Consequently, both MeanMap and InvCal completely fail, with the classification accuracy of 0\%. On the other hand, our method, which does not make strong data assumptions, achieves the perfect performance with 100\% accuracy.

\subsection{Experiments on UCI/LibSVM Datasets}
\label{subsec:uci}

\textbf{Datasets.} We compare the performance of different techniques on various datasets from the UCI repository\footnote{\url{http://archive.ics.uci.edu/ml/}} and the LibSVM collection\footnote{\url{http://www.csie.ntu.edu.tw/~cjlin/libsvmtools/}}. The details of the datasets are listed in Table \ref{table:datasets}.

In this paper, we focus on the binary classification settings.
For the datasets with multiple classes (dna and satimage), we test the one-vs-rest binary classification performance, by treating data from one class as positive, and randomly selecting same amount of data from the remaining classes as negative. For each dataset, the attributes are scaled to $[-1, 1]$.

\begin{table}[t]
\begin{center}
\begin{small}
\begin{tabular}{l|c|c|c}
\hline
Dataset & Size & Attributes & Classes \\
\hline
heart & 270 & 13 & 2 \\
\hline
heart-c & 303 & 13 & 2 \\
\hline
colic & 366 & 22 & 2 \\
\hline
vote & 435 & 16 & 2 \\
\hline
breast-cancer & 683 & 10 & 2 \\
\hline
australian & 690 & 14 & 2 \\
\hline
credit-a & 690 & 15 & 2 \\
\hline
breast-w & 699 & 9 & 2 \\
\hline
a1a & 1,605 & 119 & 2 \\
\hline
dna & 2,000 & 180 & 3 \\
\hline
satimage & 4,435 & 36 & 6 \\
\hline
cod-rna.t & 271,617 & 8 & 2 \\
\hline
\end{tabular}
\end{small}
\end{center}
\vspace{-0.2cm}
\caption{Datasets used in experiments.}
\label{table:datasets}
\vspace{-0.2cm}
\end{table}

\begin{center}
\begin{table*}[t]
{\small
\hfill{}
\scalebox{0.95}{
\begin{tabular}{l||l||c|c|c|c|c|c}
\hline
\textbf{Dataset}&\textbf{Method}& 2 & 4 & 8 & 16 & 32 & 64\\
\hline
\hline
\multirow{4}{*}{heart}   
&MeanMap         &81.85$\pm$1.28 &80.39$\pm$0.47 &79.63$\pm$0.83 &79.46$\pm$1.46 &79.00$\pm$1.42 &76.06$\pm$1.25\\
&InvCal          &81.78$\pm$0.55 &80.98$\pm$1.35 &79.45$\pm$3.07 &76.94$\pm$3.26 &73.76$\pm$2.69 &73.04$\pm$6.46\\
&alter-$\propto$SVM      &\textbf{83.41$\pm$0.71} &\textbf{81.80$\pm$1.25} &79.91$\pm$2.11 &79.69$\pm$0.64 &77.80$\pm$2.52 &76.58$\pm$2.00\\
&conv-$\propto$SVM       &83.33$\pm$0.59 &80.61$\pm$2.48 &\textbf{81.00$\pm$0.75} &\textbf{80.72$\pm$0.82} &\textbf{79.32$\pm$1.14} &\textbf{79.40$\pm$0.72}\\
\hline
\multirow{4}{*}{colic}
&MeanMap         &80.00$\pm$0.80 &76.14$\pm$1.69 &75.52$\pm$0.72 &74.17$\pm$1.61 &76.10$\pm$1.92 &76.74$\pm$6.10\\
&InvCal          &81.25$\pm$0.24 &78.82$\pm$3.24 &77.34$\pm$1.62 &74.84$\pm$4.14 &69.63$\pm$4.12 &69.47$\pm$6.06\\
&alter-$\propto$SVM      &\textbf{81.42$\pm$0.02} &\textbf{80.79$\pm$1.48} &\textbf{79.59$\pm$1.38} &\textbf{79.40$\pm$1.06} &\textbf{78.59$\pm$3.32} &\textbf{78.49$\pm$2.93}\\
&conv-$\propto$SVM       &81.42$\pm$0.02 &80.63$\pm$0.77 &78.84$\pm$1.32 &77.98$\pm$1.14 &77.49$\pm$0.66 &76.94$\pm$1.07\\
\hline
\multirow{4}{*}{vote} 
&MeanMap         &87.76$\pm$0.20 &91.90$\pm$1.89 &90.84$\pm$2.33 &88.72$\pm$1.45 &87.63$\pm$0.26 &88.42$\pm$0.80\\
&InvCal          &95.57$\pm$0.11 &95.57$\pm$0.42 &94.43$\pm$0.24 &94.00$\pm$0.61 &91.47$\pm$2.57 &91.13$\pm$1.07\\
&alter-$\propto$SVM      &\textbf{95.62$\pm$0.33} &\textbf{96.09$\pm$0.41} &\textbf{95.56$\pm$0.47} &\textbf{94.23$\pm$1.35} &\textbf{91.97$\pm$1.56} &\textbf{92.12$\pm$1.20}\\
&conv-$\propto$SVM       &91.66$\pm$0.19 &90.80$\pm$0.34 &89.55$\pm$0.25 &88.87$\pm$0.37 &88.95$\pm$0.39 &89.07$\pm$0.24\\
\hline
\multirow{4}{*}{australian}   
&MeanMap         &\textbf{86.03$\pm$0.39} &85.62$\pm$0.17 &84.08$\pm$1.36 &83.70$\pm$1.45 &83.96$\pm$1.96 &82.90$\pm$1.96\\
&InvCal          &85.42$\pm$0.28 &\textbf{85.80$\pm$0.37} &84.99$\pm$0.68 &83.14$\pm$2.54 &80.28$\pm$4.29 &80.53$\pm$6.18\\
&alter-$\propto$SVM      &85.42$\pm$0.30 &85.60$\pm$0.39 &85.49$\pm$0.78 &84.96$\pm$0.96 &85.29$\pm$0.92 &84.47$\pm$2.01\\
&conv-$\propto$SVM       &85.51$\pm$0.00 &85.54$\pm$0.08 &\textbf{85.90$\pm$0.54} &\textbf{85.67$\pm$0.24} &\textbf{85.67$\pm$0.81} &\textbf{85.47$\pm$0.89}\\
\hline
\multirow{4}{*}{dna-1}
&MeanMap         &86.38$\pm$1.33 &82.71$\pm$1.26 &79.89$\pm$1.55 &78.46$\pm$0.53 &80.20$\pm$1.44 &78.83$\pm$1.73\\
&InvCal          &93.05$\pm$1.45 &90.81$\pm$0.87 &86.27$\pm$2.43 &81.58$\pm$3.09 &78.31$\pm$3.28 &72.98$\pm$2.33\\
&alter-$\propto$SVM      &\textbf{94.93$\pm$1.05} &\textbf{94.31$\pm$0.62} &\textbf{92.86$\pm$0.78} &\textbf{90.72$\pm$1.35} &\textbf{90.84$\pm$0.52} &\textbf{89.41$\pm$0.97}\\
&conv-$\propto$SVM       &92.78$\pm$0.66 &90.08$\pm$1.18 &85.38$\pm$2.05 &84.91$\pm$2.43 &82.77$\pm$3.30 &85.66$\pm$0.20\\
\hline
\multirow{4}{*}{dna-2}
&MeanMap         &88.45$\pm$0.68 &83.06$\pm$1.68 &78.69$\pm$2.11 &79.94$\pm$5.68 &79.72$\pm$3.73 &74.73$\pm$4.26\\
&InvCal          &93.30$\pm$0.88 &90.32$\pm$1.89 &87.30$\pm$1.80 &83.17$\pm$2.18 &79.47$\pm$2.55 &76.85$\pm$3.42\\
&alter-$\propto$SVM      &\textbf{94.74$\pm$0.56} &\textbf{94.49$\pm$0.46} &\textbf{93.06$\pm$0.85} &\textbf{91.82$\pm$1.59} &\textbf{90.81$\pm$1.55} &\textbf{90.08$\pm$1.45}\\
&conv-$\propto$SVM       &94.35$\pm$1.01 &92.08$\pm$1.48 &89.72$\pm$1.26 &88.27$\pm$1.87 &87.58$\pm$1.54 &86.55$\pm$1.18\\
\hline
\multirow{4}{*}{satimage-2}
&MeanMap         &97.21$\pm$0.38 &96.27$\pm$0.77 &95.85$\pm$1.12 &94.65$\pm$0.31 &94.49$\pm$0.37 &94.52$\pm$0.28\\
&InvCal          &88.41$\pm$3.14 &94.65$\pm$0.56 &94.70$\pm$0.20 &94.49$\pm$0.31 &92.90$\pm$1.05 &93.82$\pm$0.60\\
&alter-$\propto$SVM      &\textbf{97.83$\pm$0.51} &\textbf{97.75$\pm$0.43} &\textbf{97.52$\pm$0.48} &\textbf{97.52$\pm$0.51} &\textbf{97.51$\pm$0.20} &\textbf{97.11$\pm$0.26}\\
&conv-$\propto$SVM       &96.87$\pm$0.23 &96.63$\pm$0.09 &96.40$\pm$0.22 &96.87$\pm$0.38 &96.29$\pm$0.40 &96.50$\pm$0.38\\
\hline
\end{tabular}}}
\hfill{}
\caption{Accuracy with linear kernel, with bag size 2, 4, 8, 16, 32, 64. }
\label{tb:linear}
\vspace{-0.2cm}
\end{table*}
\end{center}

\vspace{-0.7cm}
\textbf{Experimental Setup.}
Following \cite{rueping2010svm}, we first randomly split the data into bags of a fixed size.
Bag sizes of 2, 4, 8, 16, 32, 64 are tested.
We then conduct experiments with 5-fold cross validation.
The performance is evaluated based on the average classification
accuracy on the individual test instances.
We repeat the above processes 5 times (randomly selecting negative examples for the multi-class datasets, and randomly splitting the data into bags), and report the mean accuracies with standard deviations.

The parameters are tuned by an inner cross validation loop on the training subset of each partition of the 5-fold validation.
Because no instance-level labels are available during training,
we use the bag-level error on the validation bags to tune the parameters:

\begin{small}
\vspace{-0.4cm}
\begin{equation}
Err = \sum_{k=1}^T \left| \tilde{p}_k - {p}_k \right|,
\end{equation}
\vspace{-0.4cm}
\end{small}

in which $\tilde{p}_k$ and ${p}_k$ denote the predicted and the ground-truth proportions for the $k$-th validation bag. 

For MeanMap, the parameter is tuned from $\lambda \in \{0.1, 1, 10\}$.
For InvCal, the parameters are tuned from $C_p \in \{0.1, 1, 10\}$, and $\epsilon \in \{0, 0.01, 0.1\}$.
For alter-$\propto$SVM, the parameters are tuned from $C \in \{0.1, 1, 10\}$, and $C_p \in \{1, 10, 100\}$.
For conv-$\propto$SVM, the parameters are tuned from $C \in \{0.1, 1, 10\}$, and $\epsilon \in \{0, 0.01, 0.1\}$.
Two kinds of kernels are considered: linear and RBF. The parameter of the RBF kernel is tuned from $\gamma = \{0.01, 0.1, 1\}$.

We randomly initialize alter-$\propto$SVM 10 times, and pick the result with the smallest objective value. 
Empirically, the influence of random initialization to other algorithms is minimal.

\begin{center}
\begin{table}
\vspace{+0.1cm}
\renewcommand\arraystretch{1.1}
{\small
\hfill{}
\scalebox{0.9}{
\begin{tabular}{l||c|c|c}
\hline
\textbf{Method}& $2^{11}$ & $2^{12}$ & $2^{13}$\\
\hline
InvCal          &88.79$\pm$0.21 &88.20$\pm$0.62 &87.89$\pm$0.79\\
\hline
alter-$\propto$SVM      &\textbf{90.32$\pm$1.22} &\textbf{90.28$\pm$0.94} &\textbf{90.21$\pm$1.53}\\
\hline
\end{tabular}}}
\hfill{}
\caption{Accuracy on cod-rna.t, with linear kernel, with bag size $2^{11}$,  $2^{12}$, $2^{13}$.}
\label{tb:large}
\vspace{-0.4cm}
\end{table}
\end{center}

\begin{center}
\begin{table*}[t]
{\small
\hfill{}
\scalebox{0.95}{
\begin{tabular}{l||l||c|c|c|c|c|c}
\hline
\textbf{Dataset}&\textbf{Method}& 2 & 4 & 8 & 16 & 32 & 64\\
\hline
\hline
\multirow{4}{*}{heart}   
&MeanMap         &82.69$\pm$0.71 &80.80$\pm$0.97 &79.65$\pm$0.82 &79.44$\pm$1.21 &80.03$\pm$2.05 &77.26$\pm$0.85\\
&InvCal          &\textbf{83.15$\pm$0.56} &81.06$\pm$0.70 &80.26$\pm$1.32 &79.61$\pm$3.84 &76.36$\pm$3.72 &73.90$\pm$3.00\\
&alter-$\propto$SVM      &83.15$\pm$0.85 &\textbf{82.89$\pm$1.30} &\textbf{81.51$\pm$0.54} &80.07$\pm$1.21 &79.10$\pm$0.96 &78.63$\pm$1.85\\
&conv-$\propto$SVM       &82.96$\pm$0.26 &82.20$\pm$0.52 &81.38$\pm$0.53 &\textbf{81.17$\pm$0.55} &\textbf{80.94$\pm$0.86} &\textbf{78.87$\pm$1.37}\\
\hline
\multirow{4}{*}{colic}   
&MeanMap         &82.45$\pm$0.88 &81.38$\pm$1.26 &81.71$\pm$1.16 &79.94$\pm$1.33 &76.36$\pm$2.43 &77.84$\pm$1.69\\
&InvCal          &82.20$\pm$0.61 &81.20$\pm$0.87 &81.17$\pm$1.74 &78.59$\pm$2.19 &74.09$\pm$5.26 &72.81$\pm$4.80\\
&alter-$\propto$SVM      &\textbf{83.28$\pm$0.50} &\textbf{82.97$\pm$0.39} &\textbf{82.03$\pm$0.44} &\textbf{81.62$\pm$0.46} &\textbf{81.53$\pm$0.21} &\textbf{81.39$\pm$0.34}\\
&conv-$\propto$SVM       &82.74$\pm$1.15 &81.83$\pm$0.46 &79.58$\pm$0.57 &79.77$\pm$0.84 &78.22$\pm$1.19 &77.31$\pm$1.76\\
\hline
\multirow{4}{*}{vote}   
&MeanMap         &91.15$\pm$0.33 &90.52$\pm$0.62 &91.54$\pm$0.20 &90.28$\pm$1.63 &89.58$\pm$1.09 &89.38$\pm$1.33\\
&InvCal          &95.68$\pm$0.19 &94.77$\pm$0.44 &93.95$\pm$0.43 &\textbf{93.03$\pm$0.37} &87.79$\pm$1.64 &86.63$\pm$4.74\\
&alter-$\propto$SVM      &\textbf{95.80$\pm$0.20} &\textbf{95.54$\pm$0.25} &\textbf{94.88$\pm$0.94} &92.44$\pm$0.60 &\textbf{90.72$\pm$1.11} &\textbf{90.93$\pm$1.30}\\
&conv-$\propto$SVM       &92.99$\pm$0.20 &92.01$\pm$0.69 &90.57$\pm$0.68 &88.98$\pm$0.35 &88.74$\pm$0.43 &88.62$\pm$0.60\\
\hline
\multirow{4}{*}{australian}   
&MeanMap         &85.97$\pm$0.72 &85.88$\pm$0.34 &85.34$\pm$1.01 &83.36$\pm$2.04 &83.12$\pm$1.52 &80.58$\pm$5.41\\
&InvCal          &\textbf{86.06$\pm$0.30} &86.11$\pm$0.26 &\textbf{86.32$\pm$0.45} &84.13$\pm$1.62 &82.73$\pm$1.70 &81.87$\pm$3.29\\
&alter-$\propto$SVM      &85.74$\pm$0.22 &85.71$\pm$0.21 &86.26$\pm$0.61 &\textbf{85.65$\pm$0.43} &83.63$\pm$1.83 &\textbf{83.62$\pm$2.21}\\
&conv-$\propto$SVM       &85.97$\pm$0.53 &\textbf{86.46$\pm$0.23} &85.30$\pm$0.70 &84.18$\pm$0.53 &\textbf{83.69$\pm$0.78} &82.98$\pm$1.32\\
\hline
\multirow{4}{*}{dna-1}   
&MeanMap         &91.53$\pm$0.25 &90.58$\pm$0.34 &86.00$\pm$1.04 &80.77$\pm$3.69 &77.35$\pm$3.59 &68.47$\pm$4.30\\
&InvCal          &89.32$\pm$3.39 &92.73$\pm$0.53 &87.99$\pm$1.65 &81.05$\pm$3.14 &74.77$\pm$2.95 &67.75$\pm$3.86\\
&alter-$\propto$SVM      &\textbf{95.67$\pm$0.40} &\textbf{94.65$\pm$0.52} &\textbf{93.71$\pm$0.47} &\textbf{92.52$\pm$0.63} &\textbf{91.85$\pm$1.42} &\textbf{90.64$\pm$1.32}\\
&conv-$\propto$SVM       &93.36$\pm$0.53 &86.75$\pm$2.56 &81.03$\pm$3.58 &75.90$\pm$4.56 &76.92$\pm$5.91 &77.94$\pm$2.48\\
\hline
\multirow{4}{*}{dna-2}   
&MeanMap         &92.08$\pm$1.54 &91.03$\pm$0.69 &87.50$\pm$1.58 &82.21$\pm$3.08 &76.77$\pm$4.33 &72.56$\pm$5.32\\
&InvCal          &89.65$\pm$4.05 &93.12$\pm$1.37 &89.19$\pm$1.17 &83.52$\pm$2.57 &77.94$\pm$2.82 &72.64$\pm$3.89\\
&alter-$\propto$SVM      &\textbf{95.63$\pm$0.45} &\textbf{95.05$\pm$0.75} &\textbf{94.25$\pm$0.50} &\textbf{93.95$\pm$0.93} &\textbf{92.74$\pm$0.93} &\textbf{92.46$\pm$0.90}\\
&conv-$\propto$SVM       &94.06$\pm$0.86 &90.68$\pm$1.18 &87.64$\pm$0.76 &87.32$\pm$1.55 &85.74$\pm$1.03 &85.33$\pm$0.79\\
\hline
\multirow{4}{*}{satimage-2}   
&MeanMap         &97.08$\pm$0.48 &96.82$\pm$0.38 &96.50$\pm$0.43 &96.45$\pm$1.16 &95.51$\pm$0.73 &94.26$\pm$0.22\\
&InvCal          &97.53$\pm$1.33 &98.33$\pm$0.13 &98.38$\pm$0.23 &97.99$\pm$0.54 &96.27$\pm$1.15 &94.47$\pm$0.27\\
&alter-$\propto$SVM      &\textbf{98.83$\pm$0.36} &\textbf{98.69$\pm$0.37} &\textbf{98.62$\pm$0.27} &\textbf{98.72$\pm$0.37} &\textbf{98.51$\pm$0.22} &\textbf{98.25$\pm$0.41}\\
&conv-$\propto$SVM       &96.55$\pm$0.13 &96.45$\pm$0.19 &96.45$\pm$0.39 &96.14$\pm$0.49 &96.16$\pm$0.35 &95.93$\pm$0.45\\
\hline
\end{tabular}}}
\hfill{}
\caption{Accuracy with RBF kernel, with bag size 2, 4, 8, 16, 32, 64. }
\label{tb:rbf}
\end{table*}
\vspace{-0.45cm}
\end{center}

\vspace{-1.0cm}
\textbf{Results.}
Table \ref{tb:linear} and Table \ref{tb:rbf} show the results with linear kernel, and RBF kernel, respectively. 
Additional experimental results are provided in the supplementary material.
Our methods consistently outperform MeanMap and InvCal, with p-value $<$ 0.05 for most of the comparisons (more than 70\%). For larger bag sizes, the problem of learning from label
proportions becomes more challenging due to the limited amount of supervision. For these harder cases, the
gains from $\propto$SVM are typically even more significant. For instance, on
the dna-2 dataset, with RBF kernel and bag size 64, alter-$\propto$SVM outperforms
the former works by 19.82\% and 12.69\%, respectively (Table \ref{tb:rbf}).

\vspace{-0.1cm}
\textbf{A Large-Scale Experiment.}  We also conduct a large-scale
experiment on the cod-rna.t dataset containing about 271K points. The
performance of InvCal and alter-$\propto$SVM with linear kernel are
compared.  The experimental setting is the same as for the
other datasets.
The results in Table \ref{tb:large} show that
alter-$\propto$SVM consistently outperforms InvCal.  For
smaller bag sizes also, alter-$\propto$SVM outperforms InvCal, though
the improvement margin reduces due to sufficient
amount of supervision.

\vspace{+0.05cm}
\section{Discussion}

\subsection{Robustness to $\{p_k\}_{k=1}^K$}

In Section \ref{subsec:uci}, because the bags were randomly generated,
distribution of $\{p_k\}_{k=1}^K$ is approximately Gaussian for
moderate to large $K$.
It is intuitive that the performance will depend on the distribution of proportions
$\{p_k\}_{k=1}^K$.  If $p_k$ is either 0 or 1, the bags are most
informative, because this leads to the standard supervised learning
setting. On the other hand, if $p_k$'s are close to each other, the
bags will be least informative. In fact, both MeanMap and InvCal
cannot reach a numerically stable solution in such case. For
MeanMap, the linear equations for solving class means will be
ill-posed. For InvCal, because all the ``super-instances'' are assumed
to have the same regression value, the result is similar to random
guess.

$\propto$SVM, on the other hand, can achieve good performance even in
this challenging situation. For example, when using the vote dataset,
with bag sizes 8 and 32, $p_k = 38.6\%$, $\forall_{k=1}^K$(same as
prior), with linear kernel, alter-$\propto$SVM has accuracies(\%)
$94.23 \pm 1.02$ and $ 86.71\pm1.30$, and conv-$\propto$SVM has accuracies(\%) $89.60 \pm
0.59$ and $87.69 \pm 0.51$, respectively. These results are close to
those obtained for randomly generated bags in Table \ref{tb:linear}.  This
indicates that our method is less sensitive to the distribution of
$\{p_k\}_{k=1}^K$.

\vspace{-0.17cm}
\subsection{Choice of Algorithm}
Empirically, when nonlinear kernel is used, the run time of
alter-$\propto$SVM is longer than that of conv-$\propto$SVM, because
we are repeating alter-$\propto$SVM multiple times to pick the
solution with the smallest objective value.
For instance, on a machine with 4-core 2.5GHz CPU, on the vote dataset
with RBF kernel and 5-fold cross validation, the alter-$\propto$SVM
algorithm (repeating 10 times with the annealing loop, and one set of
parameters) takes 15.0 seconds on average, while the conv-$\propto$SVM
algorithm takes only 4.3 seconds.
But as shown in the experimental results, for many datasets, the performance
of conv-$\propto$SVM is marginally worse than that of alter-$\propto$SVM. This
can be explained by the multiple relaxations used in
conv-$\propto$SVM, and also the 10 time initializations of
alter-$\propto$SVM.
As a heuristic solution for speeding up the computation, one can use
conv-$\propto$SVM (or InvCal) to initialize alter-$\propto$SVM.  For
large-scale problems, in which linear SVM is used, alter-$\propto$SVM
is preferred, because its computational complexity is
$\mathcal{O}(N)$.

\vspace{-0.1cm}
The speed of both alter-$\propto$SVM and conv-$\propto$SVM
can be improved further by solving the SVM in their inner loops
incrementally. For example, one can use warm start and partial
active-set methods \cite{shilton2005incremental}. Finally, one can linearize kernels
using explicit feature maps \cite{rahimi2007random, vedaldi2012efficient},
so that alter-$\propto$SVM has linear complexity even for certain
nonlinear kernels.

\section{Conclusion and Future Work}

We have proposed the $\propto$SVM framework for learning with label
proportions, and introduced two algorithms to efficiently solve the
optimization problem. 
Experiments on several standard and
one large-scale dataset show the advantage of the proposed approach over
the existing methods. 
The simple, yet flexible form of $\propto$SVM framework naturally
spans supervised, unsupervised and semi-supervised 
learning. 
Due to the usage of latent labels, $\propto$SVM can also be potentially 
used in learning with label errors.
In the future, we will design algorithms to handle bags with 
overlapping data. Also, we plan to
investigate the theoretical conditions under which the label proportions
can be preserved with the convex relaxations.

\textbf{Acknowledgment.} We thank Novi Quadrianto and Yu-Feng Li for their help. 
We thank Jun Wang, Yadong Mu and anonymous reviewers for the insightful suggestions.

\clearpage
\bibliography{proportion_icml13_final}
\bibliographystyle{icml2013}
\end{document}